\pdfoutput=1

\documentclass[11pt]{article}

\usepackage[]{acl}

\usepackage{times}
\usepackage{latexsym}
\usepackage{amsmath}

\usepackage[T1]{fontenc}

\usepackage[utf8]{inputenc}

\usepackage{microtype}

\usepackage{inconsolata}

\usepackage{booktabs}
\usepackage{amssymb}
\usepackage{graphicx}

\newcommand{\mysubsection}[1]{\vspace{0.3em}\noindent\textbf{#1}}

\title{Corpus Considerations for Annotator Modeling and Scaling}

\author{Olufunke O. Sarumi$^{\dagger*}$ \and Béla Neuendorf$^{\dagger*}$ \and Joan Plepi$^\ddagger$ \and \\ \textbf{Lucie Flek}$^\ddagger$ \and \textbf{Jörg Schlötterer}$^{\dagger\P}$ \and \textbf{Charles Welch}$^\ddagger$\\ 
    $^\dagger$Department of Mathematics and Computer Science, University of Marburg \\ 
    $^\ddagger$Bonn-Aachen International Center for Information Technology (b-it), University of Bonn \\
    $^\P$Area Information Systems, University of Mannheim \\
    \texttt{\{sarumio,neuendob,joerg.schloetterer\}@uni-marburg.de}\\ \texttt{\{plepi,flek,cfwelch\}@bit.uni-bonn.de}
}

\begin{document}
\maketitle
\def\thefootnote{*}\footnotetext{Denotes equal contribution}\def\thefootnote{\arabic{footnote}}
\begin{abstract}
Recent trends in natural language processing research and annotation tasks affirm a paradigm shift from the traditional reliance on a single ground truth to a focus on individual perspectives, particularly in subjective tasks. In scenarios where annotation tasks are meant to encompass diversity, models that solely rely on the majority class labels may inadvertently disregard valuable minority perspectives. This oversight could result in the omission of crucial information and, in a broader context, risk disrupting the balance within larger ecosystems. As the landscape of annotator modeling unfolds with diverse representation techniques, it becomes imperative to investigate their effectiveness with the fine-grained features of the datasets in view. This study systematically explores various annotator modeling techniques and compares their performance across seven corpora. 
From our findings, we show that the commonly used user token model consistently outperforms more complex models. We introduce a composite embedding approach and show distinct differences in which model performs best as a function of the agreement with a given dataset. Our findings shed light on the relationship between corpus statistics and annotator modeling performance, which informs future work on corpus construction and perspectivist NLP.
\end{abstract}

\section{Introduction}

An integral aspect of dataset creation is obtaining multiple annotations from annotators on the same instances of data \cite{zhang2021learning,s22145245}. More often than not, these are aggregated through a majority vote to arrive at a single ground truth label~\cite{wulczyn2017ex, oro25874}. However, an annotator's background influences the label they assign. This divergence is often evident in the judgments and perceptions of annotators of subjective tasks. In the social media domain, where people's reactions are influenced by their personal experiences and vested interests, relying solely on a majority label to determine or reach a consensus proves challenging \cite{cabitza2023toward}. %
Hence, it becomes crucial to improve annotator modeling frameworks for robust user representations that capture the diverse views inherent in our datasets, while preserving individual perspectives. 

Interest in data perspectivism has been growing and with it, approaches for annotator modeling \cite{plepi-etal-2022-unifying, casola-etal-2023-confidence, davani-etal-2022-dealing}. The approaches for annotator modeling are built from corpora with unaggregated labels widely ranging in the number of annotators, data type and volume available per annotator, the type of task, and the magnitude of disagreement \cite{leonardelli-etal-2021-agreeing, kennedy2022gab, demszky2020goemotions, almanea-poesio:2022:LREC, cercas-curry-etal-2021-convabuse}. Though a few recent works have mentioned the impact of the number of annotators or the level of agreement on annotator modeling methods~\cite{kadasi2023unveiling, deng2023you, bhowmick-etal-2008-agreement}, they have not been systematically explored.

In this paper, we perform the first systematic study of the scalability of annotator modeling methods and the relationship between annotator modeling methods and corpus statistics. We implement annotator modeling and personalization techniques used in recent work \cite{mireshghallah-etal-2022-useridentifier,welch-etal-2020-exploring,plepi-etal-2022-unifying} and implement our own novel composite embedding approach.
This work sheds light on the effectiveness of annotator modeling methods under various real-world scenarios for subjective tasks. We provide recommendations for which methods to use based on the available data.

We find that when agreement is high, our composite embedding performs best, while when agreement is lower, the user token approach common in previous work performs best.
We find that the user token approach often outperforms other more complex methods developed in previous work, including the averaged SBERT, authorship attribution embeddings \cite{welch-etal-2020-compositional}, and multi-tasking \cite{davani-etal-2022-dealing}. We investigate the scalability of these methods with respect to the amount of overall data, the number of annotations per annotator, and the number of annotators in the dataset across seven different datasets and approximately 3k subsamples representing artificial datasets with controllable properties. We find that the number of annotations per annotator is the most important factor for annotator modeling, though the number of instances and annotators in the corpus both had weak but significant correlations with performance. Our code and the statistics of all trials in our experiments are publicly available\footnote{\url{https://github.com/caisa-lab/naacl2024-considerations-annotator-modeling}} to support future work that examines the relationship between corpus statistics and annotator modeling performance.

\section{Related Work}
In the first part of this section, we provide an overview of annotator modeling approaches, from modeling annotators by numeric identifiers over multitask models to representations of both, annotator and annotation information. 
In the second part, we review scaling analyses, identifying the lack of a systematic comparison between annotator modeling methods and corpus statistics.

\subsection{Annotator Modeling}
The perspectivism paradigm is fueled by the observation that  aggregate labels usually do not generalize to different demographic groups. Findings have shown that personalized models significantly enhance decision accuracy, emphasizing the substantial benefits of individualized tuning over global approaches \cite{kumar2021designing}. Likewise, disagreement within annotation processes cannot simply be dismissed as noise and merely striving for a higher inter-annotator agreement (IAA) may not always be beneficial. This calls for robust representation techniques that prioritize the inclusion of both majority and minority perspectives across the opinion spectrum~\cite{fleisig-etal-2023-majority}. 

In datasets characterized by disagreement, many methods resort to numeric identifiers for annotator modeling due to minimal annotator information availability. \citet{mireshghallah-etal-2022-useridentifier}, for instance, implemented a method where they added a non-trainable string prefix to every sentence of a user's input as a means of personalizing sentiment analysis. A simple approach is to concatenate annotator information to the instance input, which was used as a baseline by \citet{plepi-etal-2022-unifying} and \citet{deng2023you}. An improvement on this approach is seen in \citet{plepi-etal-2022-unifying} who leveraged the work of \citet{king-cook-2020-evaluating} by concatenating randomly sampled annotator comments to input text. They also computed the averaged embeddings of previous posts by annotators to represent individual annotators using a dataset of social norms; a corpus of Reddit data. 

The radical approach of training individual models per annotator~\citet{shahriar2023safewebuh} does not scale well due to computational complexity.
\citet{davani-etal-2022-dealing} reduce the computational complexity by a multi-task approach, utilizing separate fully connected layers fine-tuned for each annotator.
\citet{vitsakis2023ilab} applied this multitask architecture without further modifications in their submission to the learning with disagreement shared task \cite{leonardelli-etal-2023-semeval}. This approach, however, was only viable for datasets with few annotators such as HS-Brexit \cite{Akhtar2021WhoseOM} and ArMIS \cite{almanea-poesio:2022:LREC}, but unsuitable for MD-Agreement \cite{leonardelli-etal-2021-agreeing}. 
Similarly, \citet{sullivan-etal-2023-university} observed that the multitasking approach struggles ``to account for large or variable numbers of annotators.''

In addition to learnable annotator representations, \citet{deng2023you} trained representations for annotations. They trained compatibility matrices between the input text embedding and annotator/annotation embedding to control the influence of the latter.
They observed performance improvements in particular for tasks with high disagreement and found that demographic features are not enough to model disagreement across annotators.

\subsection{Scaling Analysis}
In this context, scaling refers to the ability of an annotator modeling approach to maintain performance and optimal throughput across varying scales of data, including the number of annotators that can be represented by a model, the number of annotators per instance and the number of annotations per annotator. Previous research has investigated the impact of a single annotation \cite{zhang2021learning} to multiple annotations per instance versus increased annotation examples \cite{Sheng2008GetAL} on performance, highlighting the importance of recognizing that there may not be a single correct interpretation for every input \cite{Aroyo2015TruthIA}. Increasing training data size for improved generalization \cite{mishra-sachdeva-2020-need} has equally been explored, giving rise to increased computational cost, which led to the use of active learning for labeling in order to manage cost \cite{fang-etal-2017-learning}. Additionally, \citet{wang2023actor} applied the multi-task model to active learning, to mitigate computational costs arising from multiple annotations. 

Few studies have explicitly examined the trade-offs between increased annotators, annotations, and training samples. While some approaches used a fixed number of annotators consistently throughout the task \cite{sullivan-etal-2023-university}, they may not be suitable for datasets where annotators vary \cite{cercas-curry-etal-2021-convabuse}. %
\citet{davani-etal-2022-dealing} excluded annotators with fewer annotations in their multi-task setup due to computational constraints.
Our study presents a systematic analysis using subsets of different corpora to assess the performance of models across varying numbers of annotators. %

\begin{table*}[t]
    \centering%
    \begin{tabular}{rllllll}
        \toprule
        & \#A & \#I & N & A/I & K-$\alpha$ & Paradigm \\\midrule
        GoEmotions & 82 & 58,012 & 2,576$\pm$ 2,292 & 3.64$\pm$ 0.94 & 0.27 & Prescriptive \\ %
        Gab Hate Speech & 18 & 27,665 & 4,807$\pm$ 3,185 & 3.13$\pm$ 0.39 & 0.25 & Prescriptive\\ %
        Social Chemistry & 2,500 & 18,431 & 46.5$\pm$ 45.9 & 6.3$\pm$ 13.3 & 0.58 & Descriptive \\ %
        MD-Agreement & 819 & 10,753 & 65.65$\pm$ 143.77 & 5.00$\pm$ 0.00 & 0.36 & Mixed \\ %
        HS-Brexit & 6 & 1,120 & 1,120.00$\pm$ 0.00 & 6.00$\pm$ 0.00 & 0.35 & Prescriptive\\ %
        ConvAbuse & 8 & 4,050 & 1,521.00$\pm$ 206.91 & 3.00$\pm$ 0.88 & 0.65 & Mixed \\ %
        ArMIS & 3 & 943 & 943.00$\pm$ 0.00 & 3.00$\pm$ 0.00 & 0.52 & Descriptive \\ %
        \bottomrule
    \end{tabular}
    \caption{Dataset statistics including the number of annotators (A), the number of total instances (I), the number of annotations per annotator (N), annotations per instance (A/I), the agreement as measured by Krippendorff's alpha, and the annotation paradigm.}
    \label{tab:all_data_stats} %
\end{table*}

\section{Datasets}\label{sec:data}

We use seven datasets from recent work on annotator modeling. All datasets use binary labels for classification. We include four datasets from the recent SemEval-2023 task on learning with disagreements~\cite{leonardelli-etal-2023-semeval}, two datasets used by \citet{davani-etal-2022-dealing}, and the Social Chemistry dataset~\cite{forbes-etal-2020-social} that was adapted for personalization and annotator modeling by \citet{plepi-etal-2022-unifying}. Dataset statistics are presented in Table~\ref{tab:all_data_stats}.

\mysubsection{Gab Hate Speech Corpus}
The Gab Hate Corpus (GHC) \citep{kennedy2022gab} comprises 27,665 social media posts from Gab.com annotated by a minimum of three annotators. The GHC features an extensive coding framework that encompasses hierarchical labels denoting dehumanizing and violent speech, markers indicating targeted groups, and rhetorical framing. As in previous annotator modeling work~\cite{davani-etal-2022-dealing}, we use labels indicating the presence or absence of hate speech. %

\mysubsection{GoEmotions}
The GoEmotions (GE) dataset \citep{demszky2020goemotions} has fine-grained emotions comprising 58k English Reddit comments and labeled for 27 emotion categories and a neutral label for no emotion. We focused on the six Ekman emotions~\cite{ekman1999basic} from the experiments of \citet{davani-etal-2022-dealing}; \emph{anger, disgust, fear, joy, sadness, and surprise}. 
Each post received annotations from three to five of a total of 82 annotators.
The agreement varies across the Ekman emotions, with Krippendorff's $\alpha$ highest at 0.35 for fear, followed by 0.29 for sadness, 0.28 for surprise, 0.27 for anger, 0.26 for joy, and 0.21 for disgust.

\mysubsection{HS-Brexit}
The Hate Speech Brexit (HSB) dataset \citep{Akhtar2021WhoseOM} is a dataset on abusive language detection consisting of 1,120 tweets related to Brexit and immigration. These were annotated for hate speech, aggressiveness, and offensiveness by two distinct groups of three annotators consisting of a target group of three Muslim immigrants in the UK and a control group of three individuals with a western background. In contrast to other datasets, the peculiarity of HS-Brexit lies in its utilization of only six annotators, distributing annotations across a smaller group with each annotator assessing many instances.

\mysubsection{ConvAbuse}
The Conversational Abuse (CVA) dataset contains approximately 4k English dialogues between users and two conversational agents~\cite{cercas-curry-etal-2021-convabuse}. Users' conversations were annotated by at least three experts in gender studies using a hierarchical labeling scheme categorized into abuse presence, abuse severity, and directness. Similarly to HS-Brexit, ConvAbuse has only a few annotators and many annotations per annotator.

\mysubsection{Multi-Domain}
The Multi-Domain (MD) Agreement dataset \cite{leonardelli-etal-2021-agreeing} comprises 10,753 tweets from three domains: Black Lives Matter (BLM), 2020 USA Presidential Elections, and COVID-19. Each tweet was annotated for offensiveness by a group of five annotators and a total of 819 annotators were recruited through Amazon Mechanical Turk (AMT).

\mysubsection{ArMIS} The Arabic Misogyny and Sexism (ArMIS) dataset contains over 900 Arabic tweets annotated specifically for the detection of misogyny and sexism~\cite{almanea-poesio:2022:LREC}. This annotation was based on identifying bias in the assessment of sexism, with a focus on the varying perspectives of the annotators concerning liberality. Three distinct individuals, self-identifying as a moderate female, liberal female, and conservative male, engaged in the annotation task. The structure of the annotation task is similar to the HS-Brexit dataset.

\mysubsection{Social Chemistry} The Social Norm (SoC) dataset~\cite{welch-etal-2022-understanding} is sourced from Reddit, an online platform with various communities known as subreddits. This dataset specifically utilizes posts and comments from the \texttt{/r/amitheasshole} (AITA) subreddit where users share social experiences and seek community opinions on the appropriateness of their behavior and that of others involved. AITA members express their views on whether the original poster is at fault in a scenario, using labels YTA (you are the asshole) and NTA (not the asshole), often providing additional reasoning. The commenters are treated as annotators and personalization methods can be applied to their comments on other parts of the subreddit or other subreddits to compute annotator representations that are not possible with most datasets that contain no more than annotator IDs (and occasionally demographic information).
The dataset comprises 21k posts and 327k verdicts (229k NTA, 98k YTA) from 86k different authors. Due to the large number of annotators, we downsampled to only the top 2,500 annotators with the highest annotation counts, resulting in 18,431 instances. 

In Table \ref{tab:all_data_stats}, we report the aggregated statistics for our datasets. These include the number of annotators (A), the number of total instances (I), the number of annotations per annotator (N), annotations per instance (A/I), and the agreement level across annotators quantified using Krippendorff’s alpha. We list the annotation paradigm in the last column, where a descriptive paradigm encourages, and a prescriptive discourages subjectivity~\cite{rottger2021two}. The MD-Agreement and ConvAbuse datasets are classified as mixed primarily because of their annotation process. While these datasets did not include a description aligned with the prescriptive paradigm and did not explicitly encourage annotator subjectivity, the guidelines allow for some degree of subjectivity.

\section{Methodology}

In our study, we investigated five distinct methods for capturing personalized attributes within a collective context, encompassing varying levels of complexity. %
Our models take the annotator ID and text that was annotated as input and are designed to distinguish between unique perspectives. In addition to representing the text in the input data, the model also incorporates a representation of the annotator who provided the annotations for the text. These representations are encoded as real-valued, low-dimensional vectors. The choice of how to represent the annotator embedding depends on the specific method being used and the information available about the annotator.
With the exception of Social Chemistry, the annotator tokens are the sole explicit distinguishing attribute of the annotators within the dataset. For several of our methods, annotator embeddings were obtained using special tokens and concatenated with the input strings for each text instance to link each text instance to its corresponding annotator. For the multi-tasking model, the annotator ID instead determines which layer of a multi-task model is responsible for predicting the label.

Additionally, we experimented with personalization techniques for the Social Chemistry dataset. This dataset contains additional writing for each annotator that can be used to derive annotator representations. We implemented the authorship attribution approach~\cite{welch-etal-2022-leveraging} and averaged SBERT embeddings~\cite{plepi-etal-2022-unifying}. Note that the averaged SBERT embeddings are averaging the additional text provided by annotators (which is only available in SoC) rather than the embeddings of the texts that were assigned an annotation.

We also introduce a \textit{composite embedding} approach, where all instances of annotations for each annotator were aggregated for each class. We calculated the average embeddings of the positive class and negative class and concatenated them to derive a unified representation for each annotator. 

\subsection{User Token Annotator Embedding}
\label{subsec:usertoken}
Consider an annotated dataset defined by $D = (X, A, Y)$ where $X$ is the set of text instances represented as $X = \{x_1, x_2, ...x_n\}$, $A$ is the set of annotators represented as $A = \{a_1, a_2, ...a_k\}$, and $Y: X \times A \rightarrow \{0,1\}$ is the annotation matrix, where an entry $y_{ij}$ represents the label assigned to instance $x_i$ by annotator $a_j$.
Since it is typical for each annotator to assign labels to only part of the dataset instances, $Y$ may contain many missing values. Note that for GoEmotions, we have multiple annotation matrices $Y_1, \dots, Y_6$, one for each of the  the six Ekman emotions.

When using a BERT-based encoder for label classification, the first step is a transformation of the individual tokens of the input text instance $x_i$ into a low-dimensional vector representation (embedding). 
We denote the input token embedding representation of instance $x_i$ as $R_i = \left[w_1, \dots, w_{|x_i|}\right]$, where $w$ is the embedding of an individual token and $|x_i|$ is the number of tokens in the input.
To incorporate annotator information, we extend the model's vocabulary and append a special token to the input; the user token. 
Each annotator is represented by a distinct user token, serving as an identifier. In accordance with input text embeddings, the user token is represented by a learnable embedding $u_j$ that is randomly initialized.
Correspondingly, the input representation for training the model augmented by annotator information becomes $R_{ij} = \left[w_1, \dots, w_{|x_i|}, u_j\right]$.
                     
\begin{table*}[t]
    \centering
    \begin{tabular}{rccccccc}
        \toprule
        Method & GE & GHC & SoC & MD & HSB & CVA & ArMIS \\
        \midrule
        SBERT & 68.6 & 68.5 & 53.3 & 73.0 & 68.6 & 85.9 & 61.7 \\
        \midrule
        User Token & \textbf{70.2} & \textbf{76.5} & 58.5 & \textbf{77.7} & \textbf{77.6} & 88.5 & 62.1 \\
        Composite Embed (Ours) & 68.2 & 68.2 & 58.6 & 73.1 & 67.6 & 85.8 & 61.4\\
        Composite+User Token (Ours) & 70.0 & 76.4 & \textbf{60.4} & 77.5 & 77.3 & \textbf{88.6} & \textbf{62.5} \\
        Multi-task & 68.3 & 70.5 & 53.5 & 75.7 & 71.7 & 82.3 & 56.6 \\ %
        \bottomrule
    \end{tabular}
    \caption{Full dataset result F1 scores on the \textbf{individual annotator} labels for each annotator representation method and dataset.} %
    \label{tab:full_results}
\end{table*}

\begin{table*}[t]
    \centering
    \begin{tabular}{rcccccc}
        \toprule
        Method & Anger & Disgust & Fear & Joy & Sadness & Surprise \\
        \midrule
        SBERT & 67.9 & 64.3 & 71.1 & 69.2 & 69.7 & 70.2 \\
        \midrule
        User Token & \textbf{69.6} & \textbf{66.9} & \textbf{73.1} & \textbf{69.9} & \textbf{70.6} & 71.1 \\
        Composite Embed (Ours) & 66.3 & 63.6 & 71.5 & 69.0 & 68.9 & 69.8 \\
        Composite+User Token (Ours) & 69.1 & 66.2 & 72.3 & 69.8 & 70.5 & \textbf{71.6} \\
        Multi-task & 67.8 & 64.0 & 70.4 & 68.9 & 68.8 & 69.6 \\
        \bottomrule
    \end{tabular}
    \caption{GoEmotions Ekman emotion F1 scores on the \textbf{individual annotator} labels for each method.}
    \label{tab:full_results_goemotions}
\end{table*}

\subsection{Composite Embedding}
\label{subsec:composite}
The composite embedding approach involves computing two embedding averages in the context of a binary classification task; the average of all instances an annotator labeled as positive, and the average of all labeled negative.
These two resulting averages represent the typical embedding patterns of the annotator when labeling positive and negative instances, respectively. 

We define the average positive embedding $E_p$ as the sum of all embeddings of all instances $x$ labeled positive by annotator $a_j$, i.e., $\{x_i|y_{ij}=1\}_{i=1}^{n}$, divided by total count of positive instances by annotator $a_j$.
\begin{equation}
    E_p = \frac{\sum_{i|y_{ij}=1} x_i^{embed}}{|\{y_{ij}=1\}_{i=1}^{n}|}
\end{equation}
We obtain the embedding of an instance $x^{embed}$ by encoding $x$ with a pre-trained SBERT model.\footnote{\url{https://huggingface.co/sentence-transformers/paraphrase-MiniLM-L6-v2}}

Similarly, let $E_n$ denote the average embedding of all instances labeled negative by annotator $a_j$, i.e., $y_{ij} = 0$:
\begin{equation}
    E_n = \frac{\sum_{i|y_{ij}=0} x_i^{embed}}{|\{y_{ij}=0\}_{i=1}^{n}|}
\end{equation}
Given $E_p$ and $E_n$, we calculate the composite embedding for annotator $a_j$ as $c_j = [E_p || E_n]$ where || denotes concatenation.

The composite embedding $c_j$ is used to initialize a special token embedding representing the annotator, whereas the user token approach $u_j$ is a random initialization. Intuitively, an initialization computed from all training data for a given annotator should provide a better starting point for the model.

\subsection{Composite Embedding with User Token}
This approach follows the convention as in \S\ref{subsec:usertoken} and \S\ref{subsec:composite} above. It utilizes the two special token embeddings; $u_j$ associated with the annotator ID of $a_j$, and $c_j$ associated with the composite representation of the same annotator. Both are appended to the input text $x_i$ annotated by $a_j$ resulting in $R_{ij} = \left[w_1, \dots, w_{|x_i|}, u_j, c_j\right]$ to model the annotator. This approach uses both a randomly initialized user token embedding and the composite embedding, which are updatable during training.

\subsection{Multi-task} The multi-tasking model is implemented on top of a BERT-base model as in previous work. One linear prediction layer is added for each annotator. The loss is summed over all annotators for a given instance.
We tuned the learning rate of the multi-tasking model on the validation sets and used $1e^{-5}$ for subsequent experiments. This model has more parameters dedicated to each annotator, so we hypothesized that it would outperform the other annotator methods. We expect to see a trade-off between the cost in time and model complexity versus the improvement in annotator modeling.

\subsection{Personalization Methods}

We implement the authorship attribution and averaged SBERT embeddings used in previous work for the Social Chemistry dataset. The averaged SBERT embeddings are computed for a given annotator by taking the set of texts that annotator has written independently of their annotations, encoding them with SBERT, and subsequently averaging the representations.

The authorship attribution method is implemented by training an authorship attribution classifier over the text of all annotators. We first embed the text with SBERT and forward these encodings to a two-layer feed-forward network. The output of the last linear layer provides a distribution over all annotators for predicting the author of the text. For each annotator, we use all of their texts from the training set (each label is accompanied with text) and pass them to the classifier. 
Intuitively, an annotator that is more often confused with another annotator should be more similar to that annotator.
    
\section{Experimental Setup}

First, following \citet{plepi-etal-2022-unifying}, we implemented a text-only baseline that also serves as the base model, which we extend by different methods for annotator modeling.
Specifically, we used SBERT \cite{DBLP:journals/corr/abs-1908-10084}, a consistently high-performing BERT-based model trained to encode sentences for a variety of downstream tasks. 
We used a pre-trained\footnote{\url{https://huggingface.co/sentence-transformers/all-distilroberta-v1}} SBERT model with the DistilRoBERTa \cite{Sanh2019DistilBERTAD} backbone, which features a 768-dimensional representation and a maximum sequence length of 512 tokens.
On top of the text encoding provided by SBERT, we implemented a classification head and fine-tuned the model for binary classification.

For the multi-task model, we used the same setup as \citet{davani-etal-2022-dealing} who used a BERT model with separate output layers for each annotator \cite{devlin-etal-2019-bert}. Lastly, since all of our datasets are in English except ArMIS, which is in Arabic, we used the Arabic BERT model from \citet{safaya-etal-2020-kuisail}, which was trained on a combination of data from Wikipedia and Common Crawl.

To evaluate, we used the macro F1 score across individual annotator's labels. We trained our models for 10 epochs, employing early stopping based on the validation set performance. The Adam optimizer was used, with an initial learning rate set at $2e^{-5}$. 
We split the data into 80\% train and 10\% for each of validation and test with the same annotators in all splits. Our experiments were conducted on a single NVIDIA A100 40GB GPU. The average running time for both training and inference phases combined was around 15 minutes per model.%

\section{Results}

\textbf{Individual Annotators} In Tables ~\ref{tab:full_results} and ~\ref{tab:full_results_goemotions} we report the F1 score for all our models across different datasets.
Our analysis reveals a consistent trend across our personalized models: performance generally improves with a smaller number of annotators and a high agreement rate. Our models are performing the worst in the Social Chemistry dataset, which contains the highest number of annotators (triple that of the dataset with the next highest count). In contrast, in the GoEmotions dataset, our models underperform compared to the MD-Agreement model, despite having fewer annotators. This can be attributed to two key differences between both datasets: a significant variance in the number of annotations per annotator and a lower overall agreement rate among annotators. Our models achieve the best performance in the ConvAbuse dataset, characterized by a smaller pool of annotators and a higher rate of agreement. This pattern suggests that both the number of annotators and the level of agreement among them are pivotal factors influencing model performance. This correlation is further observed in the GoEmotions dataset, as detailed in Table~\ref{tab:full_results_goemotions}. For instance, our models perform poorly in the \textit{disgust} category, which coincides with the lowest annotator agreement. 

For Social Chemistry, we also calculated the authorship attribution and averaged embeddings scores. The F1 for authorship attribution was 56.7 and for averaged embeddings was 57.1, both underperforming our composite embedding plus user token. However, these results are comparable with the 56 F1 reported by \citet{plepi-etal-2022-unifying} for the situation split (the same splitting method we use), even though we are using 10k fewer annotators.

\begin{table*}[t]
    \centering
    \begin{tabular}{rccccccc}
        \toprule
        Method & GE & GHC & SoC & MD & HSB & CVA & ArRMIS \\
        \midrule
        SBERT & 67.0 & 67.5 & 51.6 & 79.8 & \textbf{72.7} & 87.6 & \textbf{65.3} \\
        \midrule
        User Token & \textbf{67.4} & 69.7 & 58.7 & \textbf{80.6} & 72.2 & 88.8 & 62.1 \\
        Composite Embed (Ours) & 65.5 & 69.2 & 56.7 & 80.1 & 71.9 & 87.5 & 63.0 \\
        Composite+User Token (Ours) & 66.6 & \textbf{69.8} & \textbf{59.9} & 80.5 & 70.4 & \textbf{89.3} & 63.6 \\
        Multi-task & 66.7 & 66.7 & 49.5 & 76.4 & 70.2 & 83.1 & 55.3 \\
        \bottomrule
    \end{tabular}
    \caption{Full dataset result F1 scores for each annotator representation method and dataset using the F1 score of predicting the \textbf{majority class} label rather than the individual annotator labels.}
    \label{tab:full_results_maj}
\end{table*}

\begin{table*}[t]
    \centering
    \begin{tabular}{rcccccc}
        \toprule
        Method & Anger & Disgust & Fear & Joy & Sadness & Surprise \\
        \midrule
        SBERT & 67.5 & 62.6 & \textbf{73.2} & 63.3 & 67.5 & 67.6 \\
        \midrule
        User Token & 67.5 & 63.3 & 72.5 & 64.8 & \textbf{68.0} & \textbf{68.4} \\
        Composite Embed (Ours) & 69.4 & 61.0 & 64.9 & 63.4 & 66.9 & 67.2 \\
        Composite+User Token (Ours) & 66.8 & \textbf{63.5} & 72.2 & 64.0 & 66.2 & 66.7 \\
        Multi-task & \textbf{70.3} & 62.0 & 70.9 & \textbf{67.8} & 62.6 & 66.7 \\
        \bottomrule
    \end{tabular}
    \caption{GoEmotions Ekman emotion F1 scores for each annotator representation method using the F1 score of predicting the \textbf{majority class} label rather than the individual annotator labels.}
    \label{tab:full_results_goemotions_maj}
\end{table*}

\mysubsection{Majority Label} We presented results using the F1 scores of the individual annotator labels as our primary concern is with annotator modeling. However, it is interesting to note that the methods that perform best on annotator modeling are not the same as those that perform best when aggregating the results of individual predictions to the majority label. To obtain these results we take the individual annotator labels predicted by the model for a given instance and take the most frequent label as the predicted majority label. This is compared to the gold majority label and the resulting F1 scores are shown for the full datasets in Table~\ref{tab:full_results_maj} and for each Ekman emotion of GoEmotions in Table~\ref{tab:full_results_goemotions_maj}.

For the majority labels, we find that the composite embedding combined with the user token improves performance on GHC as well as Disgust. The performance on ArMIS still outperforms other annotator modeling methods, but the SBERT baseline appears to perform better for ArMIS, as well as HS-Brexit and Fear. For two emotions, Anger and Joy, we see the multi-task model outperform other methods, while it is never the best at predicting individual labels. For Social Chemistry, the authorship attribution method achieved an F1 of 57 and the averaged SBERT embeddings scored 56.

For the majority vote, it is interesting to note that the text-only baseline is sometimes the best model. However, we know from Tables~\ref{tab:full_results} and ~\ref{tab:full_results_goemotions} that the baseline is more often getting the individual annotator labels incorrect. This finding supports similar findings indicating that the best model of the majority class often marginalizes the voice of minority annotators~\cite{sap-etal-2019-risk,fleisig-etal-2023-majority}, which can be particularly harmful, for instance, in cases such as when racial bias impacts the perception of hate speech~\cite{sap-etal-2022-annotators}.

\section{Scaling Up}
Subsequently, to further test the impact of dataset statistics reported in Table~\ref{tab:all_data_stats}, we created subsets of our datasets by scaling the number of annotations per annotator, and the number of annotators.

When scaling the number of annotators, the three smallest datasets were scaled in increments of one annotator. For the other four, we scaled in increments of two annotators for the range of 6 to 18 annotators; the upper limit for GHC. Then for GoEmotions and Social Chemistry we scaled from 18 to 82 annotators in increments of 4. Lastly, we scaled Social Chemistry from 100 to 2.5k in increments of 100. This is repeated for each method and Ekman emotion across five runs. We tested each method on each subset, which yields a total of 1,670 trials using artificially constructed datasets.

\begin{figure}
    \centering
    \includegraphics[width=0.45\textwidth]{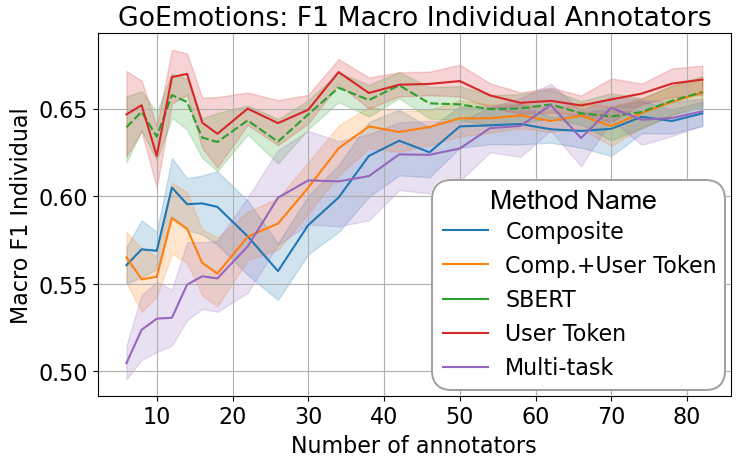}
    \caption{GoEmotions mean performance across emotions when scaling the number of annotators. The SBERT baseline is indicated by the dashed line. Shaded regions correspond to 95\% confidence intervals.}
    \label{fig:goemotions_scale_anno}
\end{figure}

Figure~\ref{fig:goemotions_scale_anno} shows each method for the GoEmotions dataset when scaling the number of annotators from 6 to 82. The baseline SBERT method is marked by a dashed line. User token performed best overall and even performs strongly when the number of annotators and amount of data is low. The composite embedding and multi-tasking methods perform poorly in this setting but approach the performance of other methods as the amount of data increases.

To identify trends across these three variables we looked at the correlation with performance across all trials.
We split the data into those with greater than 18 annotators (the median value across our corpora) and those with 18 or fewer. For each model, we calculate the percentage of relative improvement in the F1 measure over the baseline and measure the correlation using Pearson's coefficient. Since we scaled each dataset based on their available annotations and annotators to examine corpora specific patterns, this resulted in an unequal distribution of trials across datasets. For this analysis, to avoid imbalance of dataset influence, we sampled 60 trials from each of the seven corpora.

When there are more than 18 annotators, there are no significant correlations. However, when we have 18 or fewer, we find that there is a significant correlation with the dataset size ($R=0.18$, $p<0.0005$), number of annotators ($R=0.16$, $p<0.001$), and most significantly with the number of annotations ($R=0.42$, $p<0.0001$).

\begin{figure}
    \centering
    \includegraphics[width=0.48\textwidth]{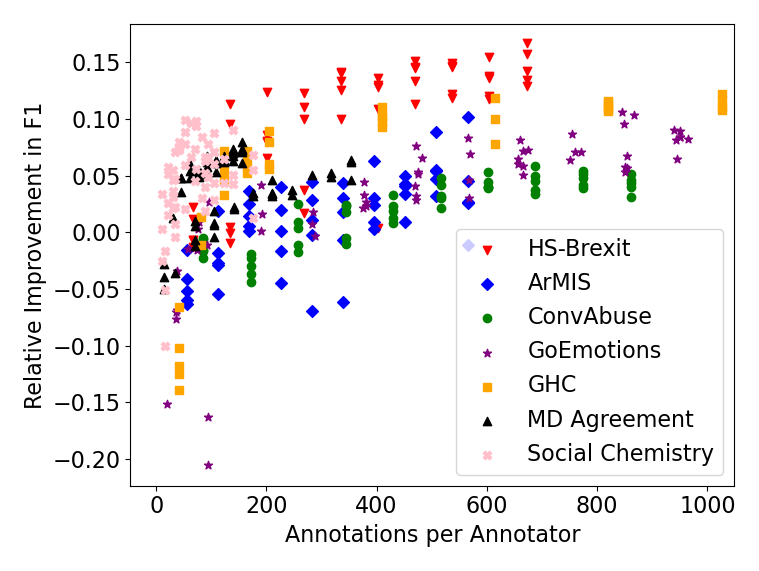}
    \caption{Relative performance increase in F1 as a function of the number of annotations per annotator.}
    \label{fig:anno_per_ann}
\end{figure}

Subsequently, we decided to look at the impact of the number of annotations per annotator on performance. We ran experiments on all datasets when fixing the number of annotators to the maximum number for ArMIS, ConvAbuse, and HS-Brexit. For the others, we fixed the number of annotators to 14 and lastly, for GoEmotions, MD Agreement, and Social Chemistry we also ran trials with 50 annotators. We varied the number of annotations per annotator in increments of 10\% up to the minimum number of annotations per annotator in each set of annotators for a given dataset. This resulted in an additional 1,260 trials, which we sampled as in the previous analysis. When examining the relationship between performance and the number of annotations per annotator, our correlation coefficient was $R=0.47$ ($p<0.0001$). 
The samples shown in Figure~\ref{fig:anno_per_ann} are from the best performing method on each dataset according to Table~\ref{tab:full_results}. 
We see that the number of annotations per annotator is important, but on most datasets the improvement levels off after a couple hundred annotations.

\section{Discussion}

When examining the dataset statistics and performance, we notice that the user token method performs well when the Krippendorff alpha is relatively low (K-$\alpha<0.4$). For the Social Chemistry, ConvAbuse, and ArMIS datasets, the agreement is higher (K-$\alpha>0.5$) and we find that in these cases the composite embedding boosts performance, becoming the best model. This may be because when agreement is higher, the composite embedding is more informative as an initialization of the annotator representation. When annotators agree more with each other, the aggregate of their positive and negative labels is more generalizable, as opposed to the random initialization of the user token.

The largest boost provided by the composite embedding occurred in the Social Chemistry dataset, where we have the lowest number of annotations per annotator. This led us to check the correlation between the performance of the user token and composite embedding methods with the number of annotations per annotator when examining the scaling experiments. We found a slightly stronger correlation for the user token ($R=0.40$) than for the composite embedding ($R=0.35$) suggesting that the low number of annotations per annotator is more detrimental to the user token method, but further work is needed to understand this relationship.

Interestingly, while the multi-task model performed the best in previous work on GoEmotions and GHC~\cite{davani-etal-2022-dealing}, we found that it performed worse in our experiments and was more expensive to train. This was the case even when using the full datasets, while our reimplementation of the approach outperforms results reported in their paper. It tends to perform worse than the baseline for the datasets that had higher agreement. Our hypothesis when we began our study was that multi-tasking would outperform other methods, as it allocates a larger number of dedicated parameters to learning representations of each annotator. We expected to see a trade-off in the model complexity and annotator modeling performance, with multi-tasking being the highest on both.%

Lastly, it is important to note that our analysis is focused on the correlations between performance and surface-level features. This is to provide initial exploratory insight into an area that deserves much further analysis and experimentation. Corpora have idiosyncrasies and future work can explore how to measure such qualities of dataset construction to support a more human-centric and personalized approach to annotator modeling, rather than one that abstracts away from these qualities.

\section{Conclusion}

As research on subjective tasks in natural language processing has grown, the importance of modeling annotators and minority opinions has become more apparent. With the recent growth in work on annotator modeling and data perspectivism, it is becoming important to understand how the properties of our datasets and methods impact the effectiveness of annotator modeling methods. We examined seven corpora to better understand recent methods and introduced our method, the composite embeddings. We found that when annotator agreement on a dataset is low (K-$\alpha<0.4$), the user token embedding was most effective. When the annotator agreement was higher (K-$\alpha>0.5$), our new composite embedding gave the best performance. Surprisingly, the user token and composite embeddings, which are simple and efficient to implement, outperformed the multi-tasking model that was the highest performing model in prior work. Importantly, we also note that the number of annotations per annotator is correlated much more strongly with performance improvements than other corpus statistics, suggesting that this should be an area of focus for those constructing new datasets and collecting annotations. Our code and collection of over 3k trial experiments statistics are publicly available to support further work in this area.

\section*{Limitations}
We examined seven different corpora for our experiments and covered the binary classification tasks of emotion recognition, hate and offensive speech detection, and judgement of social norms and morality. However, there are many more types of subjective tasks we have not considered in our work, including those that were previously thought to be more objective in nature~\cite{pavlick2019inherent}. We do not know how our results will generalize to unseen tasks that are significantly different than what we have examined in this paper. It would be interesting to extend the analysis by the type of task and the degree on a prescriptive to descriptive continuum to measure the influence on annotator modeling performance. Furthermore, our experiments examine surface statistics of corpora, which may be impacted by underlying mechanisms of data collection, corpus construction, or other factors and more work is needed to understand the relationship between these mechanisms and downstream model performance.

The scaling experiments we performed were initially designed to target the number of annotators and number of annotations per annotator and for experiments with one, we held fixed the value of the other to serve as a control. While we ran a large number of trials, many of the annotation scaling experiments use only a handful of different quantities of annotators. Future work should diversify the sampling for such trials to confirm these results, but our work serves as a first exploratory study that exemplifies the relationship with regards to the correlation and thresholds for sufficient numbers of annotations per annotator.

Our datasets did not all include demographic information. Previous work by \citet{deng2023you} included a detailed analysis and breakdown of annotator identity groups and their relation to annotator modeling performance. Since we do not know which annotators are represented in four of our seven datasets, we cannot say that our results are robust across demographic groups. Future work in this area should include more corpora with annotator information, and future data collection should strive to contextualize collected annotations with such annotator meta-data.

\section*{Acknowledgements}
This work has been supported by the Federal Ministry of Education and Research of Germany (BMBF) as a part of the Junior AI Scientists program under the reference 01-S20060, the state of North Rhine-Westphalia as part of the Lamarr Institute for Machine Learning and Artificial Intelligence, and by Hessian.AI. Any opinions, findings, conclusions, or recommendations in this material are those of the authors and do not necessarily reflect the views of the BMBF, Lamarr Institute, or Hessian.AI. We appreciate the anonymous reviewers for their detailed and constructive feedback.

\bibliography{custom}

\end{document}